\documentclass[final]{cvpr}

\usepackage{times}
\usepackage{epsfig}
\usepackage{graphicx}
\usepackage{amsmath}
\usepackage{amssymb}
\usepackage{stfloats}

\usepackage[pagebackref=true,breaklinks=true,colorlinks,bookmarks=false]{hyperref}

\def\cvprPaperID{****} 
\def\confYear{CVPR 2021}

\usepackage{url}
\usepackage{mathtools}
\usepackage[T1]{fontenc}    
\usepackage{hyperref}       
\usepackage{url}            
\usepackage{booktabs}       
\usepackage{amsfonts}       
\usepackage{nicefrac} 
\usepackage{microtype}
\usepackage{graphicx}
\usepackage{subfigure}
\usepackage{amsmath}
\usepackage{multirow}
\usepackage{array}
\usepackage{verbatim}
\usepackage{multicol}
\usepackage{datetime}
\usepackage{enumitem}
\usepackage{threeparttable}
\usepackage[ruled,vlined]{algorithm2e}

\usepackage{amsmath,amsfonts,bm}

\def\eqref#1{equation~\ref{#1}}

\def\1{\bm{1}}

\DeclareMathAlphabet{\mathsfit}{\encodingdefault}{\sfdefault}{m}{sl}
\SetMathAlphabet{\mathsfit}{bold}{\encodingdefault}{\sfdefault}{bx}{n}

\def\sS{{\mathbb{S}}}

\newcommand{\R}{\mathbb{R}}

\makeatletter
\renewcommand{\maketag@@@}[1]{\hbox{\m@th\normalsize\normalfont#1}}%
\makeatother
\newcommand*{\defeq}{\stackrel{\mathsf{def}}{=}}

\begin{document}

\title{Consistent Representation Learning for High Dimensional Data Analysis}

\author{Stan Z. Li, Lirong Wu, Zelin Zang \\
AI Lab, School of Engineering, Westlake University \& \\
Institute of Advanced Technology, Westlake Institute for Advanced Study\\
Hangzhou, Zhejiang, China \\
{\tt\small \{stan.zq.li, wulirong, zangzelin\}@westlake.edu.cn}
}

\maketitle

\begin{abstract}

High dimensional data analysis for exploration and discovery includes three fundamental tasks: dimensionality reduction, clustering, and visualization. When the three associated tasks are done separately, as is often the case thus far, inconsistencies can occur among the tasks in terms of data geometry and others. This can lead to confusing or misleading data interpretation. In this paper, we propose a novel neural network-based method, called {\em Consistent Representation Learning} (CRL), to accomplish the three associated tasks end-to-end to improve the consistencies. The CRL network consists of two nonlinear dimensionality reduction (NLDR) transformations: (1) one from the input data space to the latent feature space for {\em clustering}, and (2)  the other from the clustering space to the final 2D or 3D space for {\em visualization}. Importantly, the two NLDR transformations are performed to best satisfy {\em local geometry preserving} (LGP) constraints across the spaces or network layers, to improve data consistencies along with the processing flow. Also, we propose a novel metric, clustering-visualization inconsistency (CVI), for evaluating the inconsistencies. Extensive comparative results show that the proposed CRL neural network method outperforms the popular t-SNE and UMAP-based and other contemporary clustering and visualization algorithms in terms of evaluation metrics and visualization. 

\end{abstract}

\section{Introduction}

High dimensional data analysis for exploration and discovery includes three fundamental, associated tasks: dimensionality reduction, clustering and visualization, with the latter two rely on representations provided by dimensionality reduction procedures. When patterns of interest are non-Gaussian, a proper {\em nonlinear dimensionality reduction} (NLDR) is usually required to transform the data into latent features for {\em clustering} to work with \cite{xie2016unsupervised,yang2016joint,mcconville2019n2d}. A further NLDR may be needed to further reduce the dimensionality of the clustering feature space into two or three for {\em visualization} \cite{maaten2008visualizing,mcinnes2018umap}. 


{\bf Classic traditional clustering algorithms}, such as $K$-Means \cite{Macqueen67somemethods}, Gaussian Mixture Models (GMM) \cite{bishop2006pattern}, and spectral clustering \cite{shi2000normalized}, perform clustering based on some distance or similarity metric in the input data space. However, such measures can hardly be reliable especially for high-dimensional data, due to non-Euclidean property therein.

{\bf Clustering based on deep learning} has been proposed recently to transform the data from high-dimensional input space to a lower dimensional latent space where a distance-based metric makes more sense for clustering and visualization. For example, deep embedding clustering (DEC) \cite{xie2016unsupervised} and its variant IDEC \cite{guo2017improved} try to jointly optimize NLDR and clustering by defining a clustering-oriented objective. Instead, SpectralNet \cite{shaham2018spectralnet} embeds the input into the eigenspace of their graph Laplacian matrix and then clusters them. 

{\bf T-SNE \cite{maaten2008visualizing} and UMAP \cite{mcinnes2018umap}} are the most popular methods for combining nonlinear dimensionality reduction (NLDR) and visualization. These methods first convert Euclidean distance between data points in the input into conditional probabilities to represent the similarity and then find a lower dimensional embedding such that a cost function of KL divergence \cite{maaten2008visualizing} or cross-entropy \cite{mcinnes2018umap} is minimized.

{\bf Clustering on large image datasets} such as ImageNet and CIFAR 100, requires more sophisticated modules such as CNN, data augmentation, contrastive learning, and adversarial learning. For example, JULE \cite{yang2016joint} unifies unsupervised representation learning with clustering based on the CNN architecture to improve clustering accuracy. ASPC-DA \cite{guo2019adaptive} combines data augmentation with self-step learning to encourage the learned features to be cluster-oriented. Instead, ClusterGAN \cite{mukherjee2019clustergan} trains a generative adversarial network jointly with a clustering-specific loss to achieve clustering in the latent space. The purpose of this paper is to tackle the problem of consistent NLDR for clustering and visualization, rather than aiming at image content-based self-supervised learning or clustering.

Although NLDR, clustering and visualization are associated tasks for data analysis, they have thus far been performed independently in an uncoordinated way; namely, an NLDR is used to transform the input to a latent space in which clustering is performed, and then PCA, t-SNE, or UMAP is applied for a further NLDR to 2D for visualization. For example, the "(Not Too) Deep Clustering" (N2D) \cite{mcconville2019n2d} first uses an autoencoder \cite{hinton2006reducing} as the NLDR module to learn a lower dimensional embedding, then applies UMAP on the embedding to find the best clusterable manifolds, and finally runs $K$-Means to find the best quality clusters; for visualization, it uses another UMAP branch to further transform the embedding into 2D. In \cite{yang2016joint} and \cite{yang2019deep}, the visualization of clustering results is done by PCA and t-SNE from the latent space to 2D, respectively. In \cite{allaoui2020considerably}, UMAP is used to perform NLDR and visualization, to reduce the input data to 2D for visualization, various conventional clustering techniques are applied in the 2D visualization space. However, the dimensionality of a visualization space can be too low to provide sufficient information for clustering to achieve good results for interesting problems because the intrinsic dimensionality of manifolds can be much higher than that. 

From the above reviews on high dimensional data analysis, we find a serious problem common to the existing methods, as NLDR, clustering, and visualization are detached, the workflow not optimized in a coordinated way. {\bf Inconsistencies can occur among different stages in the processing flow}. This is the main problem we are addressing.

In this paper, we argue that the three fundamental problems of NLDR, clustering, and visualization should be solved jointly to achieve good consistencies, to improve the accuracy and interpretation for data exploration and pattern discovery in high dimensional data analysis. We propose a novel neural network-based method, called {\em Consistent Representation Learning} (CRL), to accomplish the three associated tasks in a coordinated end-to-end way and thus improve the consistencies of the representations at different stages. The CRL network consists of two NLDR transformation sub-networks: (1) one NLDR from the input layer to the latent feature layer for {\em clustering}, and (2) the other from the clustering layer to the final 2D or 3D layer for {\em visualization}, which is expected to support higher degree of nonlinearity than single-layer networks or equivalents like t-SNE or UMAP. Importantly,  {\em local geometry preserving} (LGP) constraints are imposed on the NLDR transformations across layers to best preserve the within-cluster and between-cluster structure of data to improve the consistencies of the data geometry between downstream tasks such as clustering and visualization.  Extensive comparative results show that the proposed method outperforms those methods that use or combine different NLDR methods such as autoencoder and the popular t-SNE and UMAP but not in a detached way. The performance is compared in terms of several evaluation metrics and visualization effects. The main contributions are summarized below: 

\begin{itemize}

\item[(1)] Proposing a novel neural network-based framework CRL for high dimensional data analysis, which uses two-stage LGP transformations to improve consistencies among NLDR, clustering, and visualization. The trained CRL network can generalize to unseen data.

\item[(2)] Proposing a novel metric, clustering-visualization inconsistency (CVI), for evaluating the inconsistency and explain the rationals behind.


\item[(3)] Providing better results in terms of CVI and other metrics than t-SNE and UMAP-based and other classical clustering-visualization methods.

\end{itemize}


The present work focuses on consistent representation learning of high dimensional data without using additional modules such as CNN, data augmentation, and contrastive learning. In the following, Section~2 and Section~3 introduce the CRL neural network and evaluation metrics for consistencies; Section~4 presents extensive experiments. 

\begin{figure*}[!bht]
  \centering
  \includegraphics[width=5.5in]{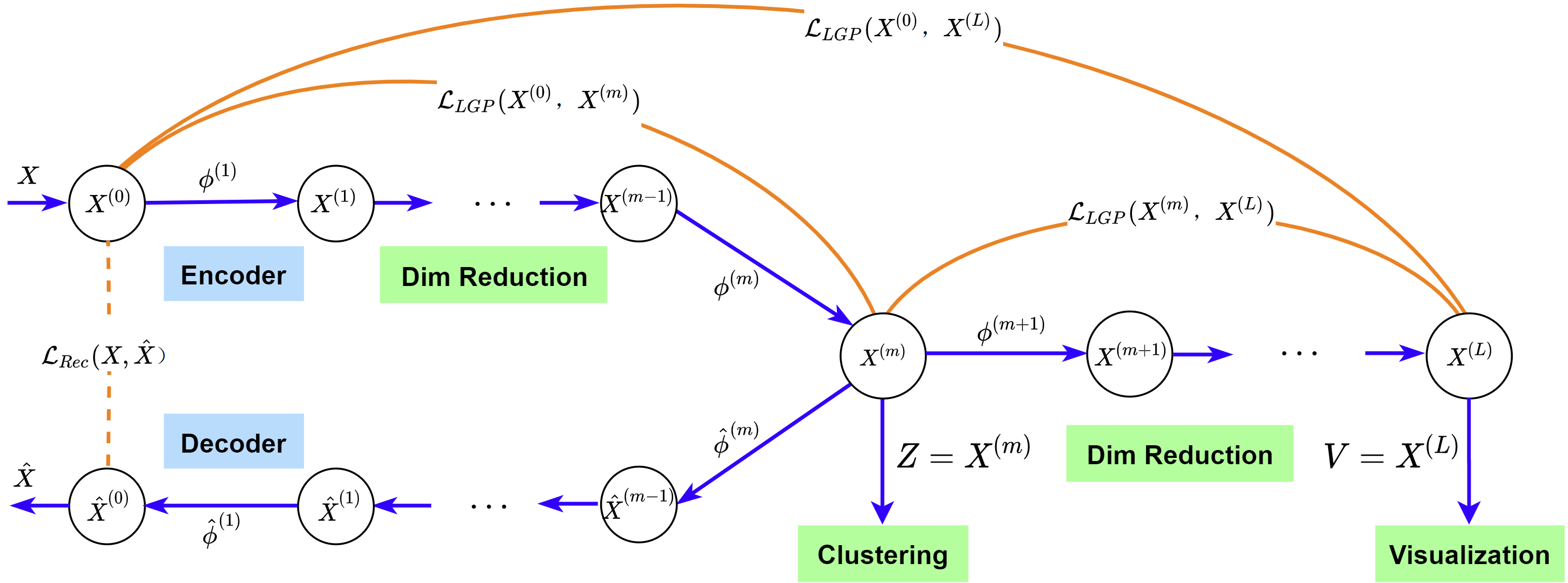}
\caption{Illustration of the CRL neural network architecture. The CRL architecture is a feed-forward neural network with a branch. It consists of two NLDR sub-networks in cascade, regularized by three cross-layer local geometry preserving (LGP) constraints. The CRL-decoder uses a reconstruction loss $\mathcal{L}_{Rec}(X,\hat{X})$ to further regularize the CRL-encoder learning. Note that the CRL-decoder starts from layer $m$ rather than layer $L$ because much of information needed for reconstruction has been lost in the 2D or 3D visualization layer. It approximates the inverse transformations of the encoder from layer 0 to layer $m$. (best viewed in color).
}
\label{fig:CRL}
\end{figure*}

\section{Consistent Representation Learning}

The basis for low dimensional representation of high dimensional data is the manifold assumption \cite{Belkin-Niyogi-02,Fefferman-manifold-2016}, which states that patterns of interest are low dimensional manifolds residing in high dimensional data space. Let $X=\left\{x_1,\ldots,x_M\right\}$ be a set of $M$ samples in the input space $\R^N$ with the index set $\sS=\left\{1,\ldots,M\right\}$. Each sample corresponds to a ground-truth label $y_{i}$ in the input space and a predicted cluster label $\hat{y}_{i}$ in the clustering space. These samples may come from one or several lower dimensional manifolds $\mathcal{M}_X\subset\mathbb{R}^N$. When $\mathcal{M}_X$ is Riemannian, its tangent subspace $T_x(\mathcal{M}_X)$ at any $x\in\mathcal{M}_X$ is locally isomorphic to an Euclidean space of dimensionality $dim(\mathcal{M}_X)<N$. 

Therefore, we can use a cascade of nonlinear neural transformations to perform the needed NLDR. The network effectively "unfolds" nonlinear manifolds in a high dimensional input space into hyper-planar regions in lower dimensional latent spaces, in a gradual way from one layer to the next. We require the transformations to be homeomorphic one after another to avoid mapping collapse. This can be implemented by applying the local geometry preserving (LGP) constraint. The LGP constraint preserves the geometric structures of data points within-cluster and between-cluster along the NLDR transformations for clustering and visualization. We use a Consistent Representation Learning (CRL) neural network to achieve these desired properties for representational consistencies among NLDR, clustering and visualization.

\subsection{CRL Neural Network} 

The CRL neural network architecture is illustrated in Fig.\ref{fig:CRL}. It is designed on the basis that the dimensionality of the clustering layer ($m$) should be higher than 2 or 3 of the visualization layer ($L$) to provide sufficient information for clustering. The architecture can be considered in two ways: (1) It is a two-stage NLDR network regularized by three cross-layer (LGP) constraints, with a branch out to a reconstruction network; (2) it is an autoencoder with a further NLDR network for visualization, regularized by the three LGP constraints. 
The LGP constraints for the two-stage NLDR network plays different roles: the Head-Mid LGP loss $\mathcal{L}_{LGP}(X^{(0)}, X^{(m)})$ between the input layer $0$ and the clustering layer $m$, the Mid-Tail LGP $\mathcal{L}_{LGP}(X^{(m)}, X^{(L)})$ for a further NLDR to the visualization layer $L$, and the Head-Tail LGP $\mathcal{L}_{LGP}(X^{(0)}, X^{(L)})$ for the overall NLDR network. The set of prescribed LGP layer pairs is denoted by $\mathbb{LP}=\{(0,m),(m,L),(0,L)\}$. The effects of these three constraints will be studied in ablation experiments. 

Following the general practice, when the dimensionality of the original data is very high, we may perform a linear dimensionality reduction, such as PCA, as a preprocessing step to obtain the input $X$ to the CRL network for further processing.

We use the Louvain algorithm \cite{blondel2008fast} as the clustering module to obtain a predicted cluster label $\hat{y}_{i}$ for each $x_i$. We choose Louvain because unlike k-means, it can cope with non-convex shaped clusters; also while a newer algorithm Leiden  \cite{traag2019louvain} is available, we found empirically that Louvain, which is more matured, produced similar quality results. Both Louvain and Leiden are clustering algorithms defined on weighted graphs. They require two preprocessing steps: (1) building a similarity affinity matrix, e.g. by using a Gaussian kernel, and (2) constructing a graph using $k$-nearest neighbor. When they perform clustering in a lower dimensional latent space, a proper NLDR with an ability to preserve geometric structure is crucial for their performance in evaluation metrics and visualized geometry. The CRL network can produce similarity matrices with the LGP property and work hand-in-hand with Louvain for clustering.

\subsection{Local Geometry-Preserving Constraint} \label{sec:LGP}

Local geometry of data at a layer can be described by similarities $u_{ij}$ computed from Euclidean distances $d(x_i,x_j)$ for all pairs of points $x_i$ and $x_j$ in $X$; on the other hand, $1-u_{ij}$ can be considered as the dissimilarity. The LGP constraint is imposed as a fuzzy divergence loss in terms of the similarities and dissimilarities between pairs across two concerned layers. The following describes the formulation.

{\bf Converting distance to similarity.} First, define the nearest neighbor (NN)-{\em normalized distance} as
$d_{i|j}\defeq d(x_i,x_j)-\rho_i \ge 0$ for $x_i\neq x_j$ where $\rho_i = d(x_i,x_{nn(i)})$ in which $x_{nn(i)}$ denotes the nearest neighbor of $x_i$. Then, $d_{i|j}$ is converted to a {\em similarity metric} $u_{i|j}(\sigma_i,\nu) = g(d_{i|j} \ | \ \sigma_i,\nu) \in (0,1)$
\begin{equation}
  g(d_{i|j} \ | \ \sigma_i,\nu)  = C_\nu  \left(1+\frac{d_{i|j}^{2}}{\sigma_{i} \ \nu } \right)^{-(\nu+1)}
  \label{equ:t2-distribution}
\end{equation}
in which $\nu\in\R_+$ is similar to the degree of freedom (DoF) parameter in the $t$-distribution and
\begin{equation}
C_\nu  = 2\pi \left(\frac{\Gamma\left(\frac{\nu+1}{2}\right)}
{\sqrt{\nu \pi} \Gamma\left(\frac{\nu}{2}\right)}\right)^2
\label{equ:t2-nu}
\end{equation}
is a function of $\nu$ which sets the limit $\lim_{\nu\to +\infty} g(0 \ | \ \sigma_i, \nu)=1$ ($\forall \sigma_i>0)$).  The data-calibration parameter $\sigma_i>0$ is estimated from the data by binary search to best fit to the following the equation
\begin{equation}
\sum_{j\neq i} u_{i|j}(\sigma_i,\nu)=\log_2 Q
\label{equ:sigma_i}
\end{equation}
for the perplexity-like hyperparameter $Q$ given. We find empirically Equ.~(\ref{equ:t2-distribution}) a slightly better choice for distance-similarity conversion than others, such as the normal function and Cauchy function in t-SNE \cite{maaten2008visualizing} and the fitted polynomial function in UMAP \cite{mcinnes2018umap}. Finally, the following symmetrization is performed $u_{ij}=u_{j  |  i}+u_{i | j} - u_{j  |  i} u_{i | j}$. Note that $u_{ij}$ are confined in the range $(0,1)$ to be a fuzzy set but their sum needs not to be 1 to be a probability distribution.

{\bf LGP loss.} A LGP loss between layers $l$ and $l'$ is defined in terms of similarities computed at the resepective layers:
\begin{footnotesize}
\begin{equation}
  \begin{aligned}
  \hspace{-2mm}
    \mathcal{L}_{LGP}^{(l,l')}(W) = 
    \sum_{i,j \in \sS, i \neq j}
    u^{(l)}_{i j} \log \frac{u^{(l)}_{i j}}{u^{(l')}_{i j}}
    + (1-u^{(l)}_{i j}) 
    \log 
    \frac{ 1-u^{(l)}_{i j}} {1-u^{(l^{'})}_{i j}}
  \end{aligned} 
  \label{equ:elis-loss}
\end{equation}
\end{footnotesize}
\noindent The above formulation is also known as the "fuzzy information for discrimination" in  \cite{Bhandari-IS-1993} and the "fuzzy set cross entropy" in UMAP \cite{mcinnes2018umap}. However, unlike t-SNE and UMAP, which minimize losses w.r.t. the embedding, the CRL network minimizes $\mathcal{L}_{LGP}$ w.r.t. the transformation matrix $W$. It encourages minimizing the sum of the differences between similarities $\log u^{(l)}_{i j}$ and $\log u^{(l')}_{i j}$ weighted by $u^{(l)}_{i j}$, and also the weighted differences between the dissimilarities, thus preserving the geometric structure.

\subsection{CRL Encoder and Decoder}

\textbf{The CRL-encoder} consists of two manifold-based NLDR transformation networks in cascade under the LGP constraints, as shown in Fig.~\ref{fig:CRL}. A CRL-encoder is able to learn an NLDR transformation without the need for a decoder (unlike an autoencoder), and the learned encoder can generalize to unseen data whereas that t-SNE and UMAP cannot. The total loss for the CRL-encoder is 
\begin{equation}
\mathcal{L}_{Enc}(W)=\sum_{(l,l')\in\mathbb{LP}} \alpha^{(l,l')} L_{LGP}^{(l,l')}(W)
\label{equ:loss-elis-enc}
\end{equation}
where $\alpha^{(l,l')}$ are the weights. 

\textbf{The CRL-decoder} starts from layer $m$ (clustering space) to reconstruct $X$ (see Fig.~\ref{fig:CRL}) and has a structure symmetric to layers $0$ to $m$ of the CRL-encoder. It approximates the inverse transformation and learned with the reconstruction loss  $\mathcal{L}_{Rec}(W)= \sum_{i\in\sS} \parallel{x_i}-\hat{x}_i\parallel^2$. Experiments show that $\mathcal{L}_{Rec}(W)$ is beneficial to improve visualization and clustering results.


\section{Evaluation Metrics}

We are interested in evaluating label inconsistency between samples in the input and clustering spaces and inconsistency between those in the clustering and visualization spaces. Two most commonly used metrics are the accuracy (ACC) and normalized mutual information (NMI) for the evaluation of the inconsistency between the {\em ground-truth labels} of the input data and the {\em predicted cluster labels} obtained by clustering. However, ACC and NMI cannot measure the inconsistency between the  {\em predicted cluster labels} and {\em ground-truth labels} in the visualization embedding. To the best of our knowledge, there exists no metric for measuring such inconsistency. Therefore, we will define a novel metric, called \textit{Clustering-Visualization Inconsistency} (CVI) for this purpose. 

{\bf  Accuracy (ACC)}  is a performance metric specifically designed to measure the accuracy of unsupervised clustering  \cite{mcconville2019n2d,guo2017improved,shaham2018spectralnet}. To use the ACC metric, a mappings $\psi$ from  predicted cluster labels to the ground-truth labels has to be computed by the following formula 
\begin{equation}
A C C=\max _{\mathcal \psi} \frac{\sum_{i\in\sS} \ell\left\{y_{i}=\mathcal \psi\left(\hat{y}_{i}\right)\right\}}{M}
\end{equation}
\noindent where $y_i$ and $\hat{y}_i$ are the ground-truth label and the predicted cluster label for $x_i$, respectively. $\mathcal \psi(\cdot)$ is maximized over all one-to-one mappings from one label domain to the other using Hungarian algorithm \cite{kuhn1955hungarian}.

{\bf Normalized Mutual Information (NMI)} is an information theoretic measure of how close the predicted cluster labels $\hat{y}$ are to the ground-truth labels $y$  \cite{yang2016joint,mcconville2019n2d,xie2016unsupervised}. It is defined as
\begin{equation}
N M I=\frac{I(\hat{y} ; y)}{\max \{H(\hat{y}), H(y)\}}
\end{equation}
where $I(\hat{y}; y)$ is the mutual information between the ground-truth labels $y$ and the predicted cluster labels $\hat{y}$, and $H(\cdot)$ denotes their entropies. NMI is in the range of [0, 1], with 0 meaning no correlation and 1 exhibiting perfect correlation.

{\bf Clustering-Visualization Inconsistency (CVI)} is defined as the total distance between misclustered points and the centers of the cluster they actually belong to, as follows
\begin{equation}
C V I =\frac{1}{M}\sum_{i\in\sS} \frac{||v_i-C(y_i)||_2\cdot\ell\left\{\hat{y}_{i} \ne y_{i}\right\}}{||\hat{C}(\hat{y}_i)-C(y_i)||_2}
\end{equation}
where $v_i$ is the embeddings of $x_i$ in the visualization space, $y_i$ is the ground-truth label, $\hat{y}_i$ is cluster label assignment obtained by Hungarian algorithm , and $\ell(\cdot)$ is an indicator function. $||\hat{C}(\hat{y}_i)-C(y_i)||_2$ is a scale normalization term, making the CVI metric scale-independent; $C(y_i)$ denotes the center of all data points with {\em ground-truth label} $y_i$ in the embedding, defined as
\begin{equation}
C(y_i)=\frac{\sum_{j\in\sS} v_j\cdot\ell\left\{y_{j} =y_{i}\right\}}{\sum_{j\in\sS} \ell\left\{y_{j} =y_{i}\right\}}
\end{equation}
and similarly, $\hat{C}(\hat{y}_i)$ denotes the center of all data points with {\em cluster label assignment} $\hat{y}_i$ in the embedding, defined as
\begin{equation}
\hat{C}(\hat{y}_i)=\frac{\sum_{j\in\sS} v_j\cdot\ell\left\{\hat{y}_{j} =\hat{y}_{i}\right\}}{\sum_{j\in\sS} \ell\left\{\hat{y}_{j} =\hat{y}_{i}\right\}}
\end{equation}
A larger CVI value indicates more inconsistency between clustering and visualization. Note that CVI can be generalized to measure inconsistency between corresponding labels of transformed entities in any two spaces other than clustering and visualization spaces.

\section{Experiments}


\subsection{Settings}


The proposed CRL method is evaluated in comparison t-SNE and UMAP-based clustering-visualization algorithms and several state-of-the-art clustering (but without visualization) techniques, including JULE \cite{yang2016joint}, N2D \cite{mcconville2019n2d} and ASPC-DA \cite{guo2019adaptive}. {\bf Five datasets} are selected to evaluate clustering and visualization performance: MNIST-full, MNIST-test, USPS, Fashion-MNIST, and Coil-100. The number of samples, categories, and data size are listed in Table \ref{tab:dataset}. MNIST-full \cite{lecun1998gradient}: A dataset consists of 60000 training samples and 10000 test samples, each being a 28$\times$28 handwritten digit image. MNIST-test: A dataset containing only the testing part of MNIST-full. USPS: A handwritten digit dataset composed of 9298 samples, with a size of 16$\times$16 pixels. Fashion-MNIST \cite{xiao2017fashion}: The dataset has the same number and size of images as MNIST-full, but it consists of various types of fashion products rather than numbers. Coil-100 \cite{nene1996columbia100}: A dataset containing 7200 samples from 100 objects at 72 different angles.

\vspace{-5mm}
\begin{table}[htbp]
	\begin{center}
	\caption{Datasets used in our experiments.}
	\label{tab:dataset}
	\begin{tabular}{lccc}
		\hline
		\multicolumn{1}{l}{\textbf{Dataset}} & \multicolumn{1}{c}{\textbf{Samples}} & \multicolumn{1}{c}{\textbf{Categories}} & \multicolumn{1}{c}{\textbf{Data Size}} \\ \hline
		MNIST-full     & 70000 & 10  & 28$\times$28$\times$1 \\
		MNIST-test     & 10000 & 10  & 28$\times$28$\times$1 \\
		USPS           & 9298  & 10  & 16$\times$16$\times$1 \\
		Fasion-Mnist   & 70000 & 10  & 28$\times$28$\times$1 \\
		Coil100        & 7200  & 100 & 55$\times$55$\times$3 \\\hline
	\end{tabular}
	\end{center}
\end{table}
\vspace{-5mm}


{\bf The CRL-encoder structure} is $d$-1000-500-300-100-50-30-2, where $d$, \textit{100}, and \textit{2} are the dimensions of the input, clustering space and visualization space. The self-reconstruction is performed from the clustering space (rather than the visualization space), and the CRL-decoder structure is 100-300-500-100-$d$. The LeakyReLU is used as the activation function. Louvain with optimal resolution (resolution that makes the number of resulting clusters equal to true) is used for clustering. The $k$-nearest neighbor ($k$NN) with $k$ = 10 ($k$ = 20 for Coil-100) is used to build graphs in the clustering space for Louvain. The following parameters are set the same for all datasets: Adam optimizer \cite{kingma2014adam} with learning rate $lr$ = 0.001; epochs $E$ = 1500; parameter $\nu$ for the input, clustering, visualization space: $\nu_{i}$ = 100, $\nu_{c}$ = 0.001, $\nu_{v}$: 0.001$\rightarrow$100; loss weights $\alpha^{(0, L)}$ = 1.0, $\alpha^{(0, m)}$ = $\alpha^{(m, L)}$ = 0.5 (0.1 for Coil-100), $\alpha^{(0, 0')}$ = 0.1. Due to page limitations, some dataset-specific hyperparameters ($Q$ and batchsize) are summarized in Table \ref{tab:hyperparameter} in \textbf{Appendix A1}. The parameters $\nu_{v}$, $Q$, Louvain resolution can be used as user-defined parameters to provide different clustering and visualization effects. The rest of the parameters can be fixed in practice. The implementation uses the PyTorch 1.6.0 library running on Ubuntu 18.04 on NVIDIA v100 GPU. On the MNIST-full dataset with 70000 data points, it takes about 2.5 hours to run 1500 epochs.

\begin{table*}[htbp]
\begin{center}
\caption{Clustering performance of t-SNE and UMAP-based clustering-visualization algorithms on 5 datasets.}
\label{tab:comparision}
\begin{tabular}{llccccccc}
\hline
 &  & Raw Data & AE & t-SNE & UMAP & AE-TSNE & AE-UMAP & CRL \\ \hline
\multirow{3}{*}{MNIST-test} & ACC & 0.799 & 0.646 & 0.834 & 0.846 & 0.857 & 0.947 & \textbf{0.950}  \\
 & NMI & 0.814 & 0.599 & 0.825 & 0.842 & 0.860 & 0.883 & \textbf{0.885}  \\
 & CVI & 0.155 & 0.139 & 0.259 & 0.205 & 0.103 & 0.054 & \textbf{0.050}  \\ \hline
\multirow{3}{*}{MNIST-full} & ACC & 0.826 & 0.601 & 0.955 & 0.874 & 0.886 & 0.887 & \textbf{0.976} \\
 & NMI & 0.867 & 0.654 & 0.899 & 0.890 & 0.920 & 0.923 & \textbf{0.932} \\
 & CVI & 0.139 & 0.159 & 0.040 & 0.112 & 0.116 & 0.082 & \textbf{0.022} \\ \hline
\multirow{3}{*}{USPS} & ACC & 0.959 & 0.611 & 0.892 & 0.961 & 0.893 & 0.960 & \textbf{0.962} \\
 & NMI & 0.904 & 0.669 & 0.895 & 0.908 & 0.889 & 0.904 & \textbf{0.918} \\
 & CVI & 0.037 & 0.347 & 0.088 & 0.037 & 0.083 & 0.041 & \textbf{0.034} \\ \hline
\multirow{3}{*}{Fashion-MNIST} & ACC & 0.616 & 0.499 & 0.502 & 0.570 & 0.603 & 0.603 & \textbf{0.653} \\
 & NMI & 0.675 & 0.541 & 0.643 & 0.626 & 0.668 & 0.680 & \textbf{0.691} \\
 & CVI & 0.390 & 0.509 & 0.602 & 0.428 & 0.461 & 0.420 & \textbf{0.329} \\ \hline
\multirow{3}{*}{Coil100} & ACC & 0.686 & 0.656 & 0.775 & 0.760 & 0.823 & 0.837 & \textbf{0.949}  \\
 & NMI & 0.858 & 0.834 & 0.925 & 0.898 & 0.940 & 0.931 & \textbf{0.980}  \\
 & CVI & 0.137 & 0.136 & 0.302 & 0.248 & 0.443 & 0.154 & \textbf{0.063}  \\ \hline
\end{tabular}
\end{center}
\end{table*}

\begin{table*}[bp]
\begin{center}
\caption{Clustering performance of several classical clustering (but without visualization) algorithms on 5 datasets.}
\label{tab:sota}
\begin{tabular}{lcccccccccc}
\hline
\multirow{2}{*}{} & \multicolumn{2}{c}{MNIST-test} & \multicolumn{2}{c}{MNIST-full} & \multicolumn{2}{c}{USPS} & \multicolumn{2}{c}{Fashion-MNIST} & \multicolumn{2}{c}{Coil100} \\
 & ACC & NMI & ACC & NMI & ACC & NMI & ACC & NMI & ACC & NMI \\ \hline
K-Means \cite{Macqueen67somemethods} & 0.546 & 0.501 & 0.532 & 0.500 & 0.668 & 0.601 & 0.474 & 0.512 & 0.543* & 0.819* \\
SC-Nut \cite{shi2000normalized} & 0.660 & 0.704 & 0.656 & 0.731 & 0.649 & 0.794 & 0.508 & 0.575 & 0.632* & 0.827*  \\
GMM \cite{bishop2006pattern} & 0.464 & 0.465 & 0.389 & 0.333 & 0.562 & 0.540 & 0.463 & 0.514 & 0.433* & 0.728* \\ 
DEC \cite{xie2016unsupervised} & 0.856 & 0.830 & 0.863 & 0.834 & 0.762 & 0.767 & 0.518 & 0.546 & 0.489* & 0.767*  \\
IDEC \cite{guo2017improved} & 0.846 & 0.802 & 0.881 & 0.867 & 0.761 & 0.785 & 0.529 & 0.557 & 0.565* & 0.801* \\
DeepCluster \cite{caron2018deep} & 0.854 & 0.713 & 0.797 & 0.661 & 0.562 & 0.540 & 0.542 & 0.510 & - & - \\
VaDE \cite{jiang2016variational} & 0.287 & 0.287 & 0.945 & 0.876 & 0.566 & 0.512 & 0.578 & 0.630 & - & - \\
SpectralNet \cite{shaham2018spectralnet} & 0.817 & 0.821 & 0.800 & 0.814 & 0.834* & 0.818* & 0.508* & 0.529* & 0.654* & 0.841* \\
JULE \cite{yang2016joint} & \underline{0.961} & \underline{0.915} & 0.964 & 0.913 & 0.950 & \textit{0.913} & 0.563 & 0.608 & \textit{0.908}* & \textit{0.939}* \\
ClusterGAN \cite{mukherjee2019clustergan} & - & - & 0.950 & 0.890 & - & - & \textit{0.630} & 0.640 & - & - \\
ASPC-DA \cite{guo2019adaptive} & \textbf{0.973} & \textbf{0.936} & \textbf{0.988} & \textbf{0.966} & \textbf{0.982} & \textbf{0.951} & 0.591 & \textit{0.654} & \underline{0.914}* & \underline{0.948}* \\
N2D \cite{mcconville2019n2d} & 0.948 & 0.882 & \underline{0.979} & \underline{0.942} & \textit{0.958} & 0.901 & \textbf{0.672} & \underline{0.684} & 0.847* & 0.935*  \\
CRL & \textit{0.950} & \textit{0.885} & \textit{0.976} & \textit{0.932} & \underline{0.962}& \underline{0.918} & \underline{0.653} & \textbf{0.691} & \textbf{0.949} & \textbf{0.980} \\ \hline
\end{tabular}
\end{center}
\end{table*}

\subsection{Consistency of Input and Clustering}

Because of the non-Euclidean nature of the high dimensional data in the input space, the inconsistency between the input and clustering spaces may be less intuitive. Therefore, we evaluate the consistency between the two spaces using ACC and NMI, which measure the consistency between true labels in the input space and predicted labels obtained by clustering.

The existing methods that involve NLDR, visualization and clustering, generally perform dimensionality reduction via autoencoder (AE) and visualization with t-SNE and UMAP, which are two completely independent stages. To compare with this pipeline, we perform clustering in the 2D embeddings obtained by AE, t-SNE, and UMAP. Additionally, the best clusterable manifolds on the 10D (100D for Coil-100) embeddings (learned using an autoencoder) are determined by t-SNE and UMAP, and then clustered. We denote these two methods as AE-TSNE and AE-UMAP, and include them in the comparison. Finally, the results of direct clustering on the Raw Data are provided as a baseline. 

The metrics ACC/NMI/CVI of the above methods are reported in Table \ref{tab:comparision}. The CRL outperforms the AE by a significant margin and surpasses the other five compared methods on all datasets. In particular, we obtain the best performance on the Coil-100 dataset, and more notably, our clustering accuracy exceeds the second by 11.2\%, demonstrating a significant advantage on complex datasets. Furthermore, while good visualizations can sometimes be obtained using t-SNE and UMAP, their clustering accuracy is not always high, e.g., on the Fashion-MNIST dataset even inferior to the clustering performance on the Raw Data. This suggests that applying embeddings obtained with t-SNE and UMAP directly to clustering can easily lead to inconsistencies between clustering and visualization. We will discuss the clustering-visualization inconsistency in detail in Sec.~\ref{sec:consistency}.

The metrics ACC/NMI of several classical clustering  algorithms (without visualization) are reported in Table \ref{tab:sota} where the \textbf{bold}, \underline{underline}  and  \textit{italic} represent the top results of rank 1, 2 and 3, respectively. For those comparison methods whose results are not reported on some datasets, we run the released code using the hyperparameters provided in their paper and label them with (*). When the code is not publicly available, or running the released code is not practical, we put dash marks (-) instead of the corresponding results. The selected clustering algorithms include some conventional non-deep methods, such as $K$-means \cite{Macqueen67somemethods}, SC-Nut \cite{shi2000normalized}, and GMM \cite{bishop2006pattern}, as well as more recent deep clustering methods such as DEC \cite{xie2016unsupervised}, JULE \cite{yang2016joint}, ASPC-DA \cite{guo2019adaptive} and N2D \cite{mcconville2019n2d}, etc. Among all methods, only CRL unifies NLDR, clustering and visualization into a single framework that enables end-to-end training. Other methods, however, are limited to clustering and rely on other techniques, such as t-SNE and UMAP, to visualize their clustering results. Since comparisons with t-SNE and UMAP on CVI metric are already provided in Table \ref{tab:comparision}, it will not be repeated here.

\begin{figure*}[bp]
	\begin{center}
		\includegraphics[width=0.95\linewidth]{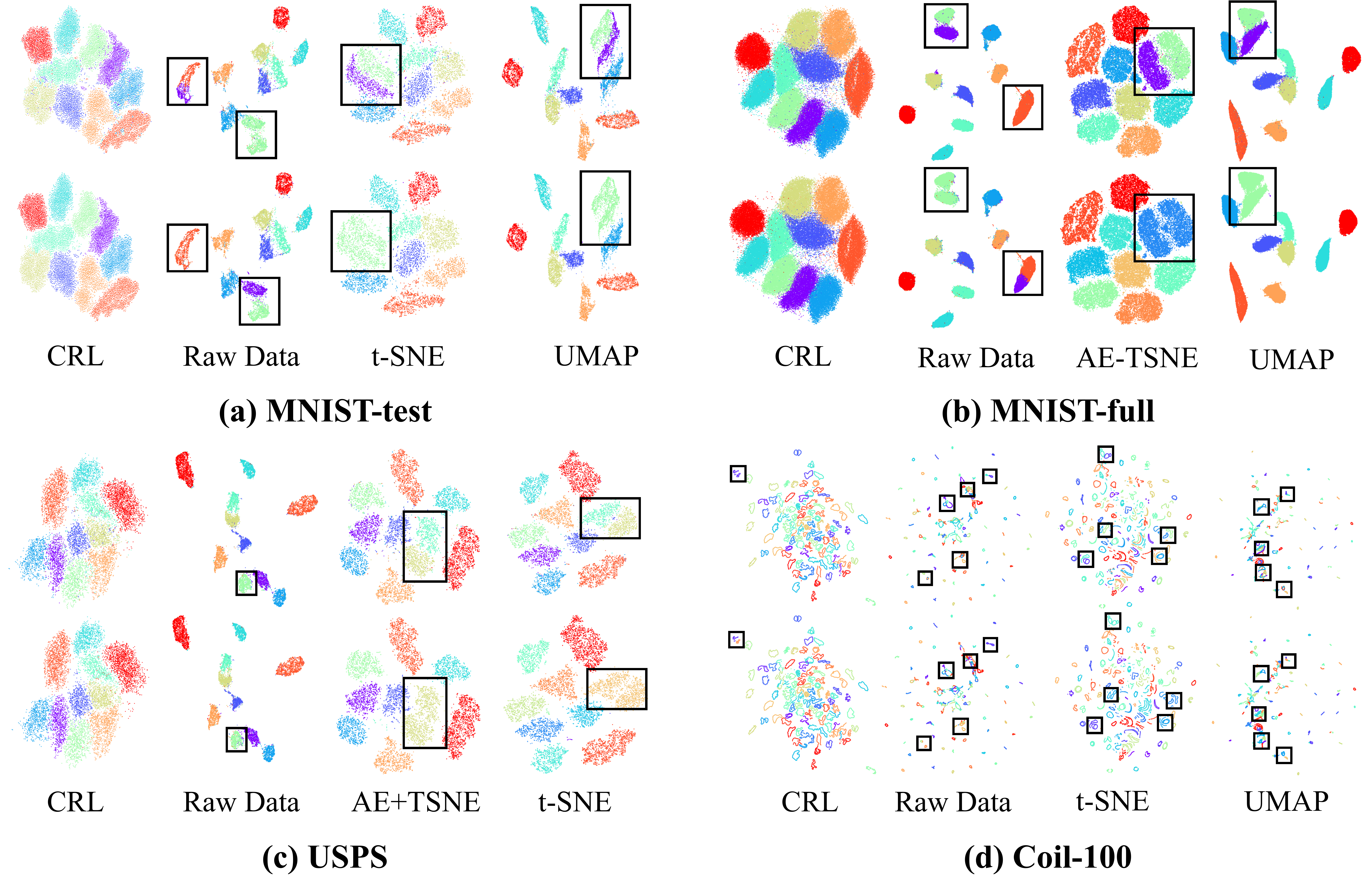}
	\end{center}
	\caption{Inconsistency analysis for visualization and clustering. The visualizations in the subplots are marked by true labels (upper row) and clustered predicted labels (lower row), respectively.}
	\label{fig:consistency}
\end{figure*}


As shown in Table \ref{tab:sota}, for the 10 performance numbers on the five datasets, CRL has 3 results in the first place, 3 results in the second, and 4 results in the third. While ASPC-DA achieves the best performance on three datasets (MNIST-test, MNIST-full, and USPS), its performance gains do not come directly from clustering, but from sophisticated modules such as data augmentation and self-paced learning. Once these modules are removed, there is a very large degradation in its performance. For example, with data augmentation removed, ASPC-DA achieves less competitive performance, e.g., an accuracy of 0.924 (\textit{vs} 0.988) on MNIST-test, 0.785 (\textit{vs} 0.973) on MNIST-full and 0.688 (\textit{vs} 0.982) on USPS. However, this paper focuses on the consistency among NLDR, clustering, and visualization and has not used those sophisticated modules for better performance. Under a fairer comparison (without taking ASPC-DA into account), CRL performs the best overall for the 10 performance numbers on the five datasets. In particular, on the Coil-100 dataset, CRL achieves the best performance for both ACC and NMI (even better than ASPC-DA).

\subsection{Consistency of Clustering and Visualization} \label{sec:consistency}

The CVI metric measures inconsistency by the total distance between misclustered points and the centers of the cluster they actually belong to in the visualization. From the CVI metric in Table \ref{tab:comparision}, CRL achieves the best performance on all datasets. More notably on the USPS dataset, though CRL is only 0.002 more accurate than AE-UMAP, its CVI metric is 17.7\% lower, demonstrating the great advantage of LGP constraint-based CRL in improving consistency.

Additionally, we present the visualization results for CRL and other compared methods on 4 datasets in Fig.~\ref{fig:consistency}, marked with true and clustered prediction labels, respectively. The Fashion-MNSIST dataset is not included in the comparison because all methods are poorly visualized on this dataset, making it difficult to analyze inconsistencies. Besides, we have not shown results for all corresponding methods, because not all methods suffer from inconsistency problem across all datasets, so we present only a few representative results for analysis. As boxed in Fig.~\ref{fig:consistency}, samples from one category in the visualization can be clustered into two clusters, while samples from different categories in the visualization can be clustered into the same cluster by methods other than CRL. Both types of inconsistencies are prevalent across the four datasets. In addition, some methods, such as Raw Data's results on the USPS dataset (visualized by UMAP), fail to identify noisy points in the embedding; instead, clustering even makes the visualization noisier, which is also inconsistent. Furthermore, we note that inconsistencies are more severe on the Coil-100 dataset, with a large number of misclustered samples at the boundaries of different clusters.

\begin{table*}[tbp]
\begin{center}
\caption{Clustering performance for ablation study on 5 datasets.}
\label{tab:ablation}
\begin{tabular}{llccccccc}
\hline
 &  & Scheme A & Scheme B & Scheme C & Scheme D & Scheme E & Scheme F & Scheme G \\ \hline
\multirow{3}{*}{MNIST-test} & ACC & 0.817 & 0.709 & 0.938 & 0.934 & 0.900 & 0.686 & \textbf{0.950}  \\
 & NMI & 0.835 & 0.814 & 0.867 & 0.860 & 0.848 & 0.801 & \textbf{0.885} \\
 & CVI & 0.168 & 0.331 & 0.054 & 0.086 & 0.093 & 0.354 & \textbf{0.050} \\ \hline
\multirow{3}{*}{MNIST-full} & ACC & 0.930 & 0.972 & 0.847 & 0.837 & 0.969 & 0.785 & \textbf{0.976} \\
 & NMI & 0.915 & 0.927 & 0.884 & 0.889 & 0.920 & 0.872 & \textbf{0.932} \\
 & CVI & 0.071 & 0.024 & 0.669 & 0.795 & 0.043 & 0.821 & \textbf{0.022} \\ \hline
\multirow{3}{*}{USPS} & ACC & 0.951 & 0.846 & 0.957 & 0.950 & 0.823 & 0.832 & \textbf{0.962} \\
 & NMI & 0.891 & 0.877 & 0.903 & 0.891 & 0.854 & 0.858 & \textbf{0.918} \\
 & CVI & 0.042 & 0.193 & 0.042 & 0.045 & 0.164 & 0.213 & \textbf{0.034} \\ \hline
\multirow{3}{*}{Fashion-MNIST} & ACC & 0.636 & 0.584 & 0.629 & 0.623 & 0.649 & 0.624 & \textbf{0.653} \\
 & NMI & 0.673 & 0.645 & 0.688 & 0.693 & 0.660 & 0.683 & \textbf{0.691} \\
 & CVI & 0.351 & 0.395 & 0.830 & 1.100 & 0.404 & 0.748 & \textbf{0.329} \\ \hline
\multirow{3}{*}{Coil100} & ACC & 0.907 & 0.848 & 0.911 & 0.889 & 0.934 & 0.790 & \textbf{0.949}  \\
 & NMI & 0.962 & 0.963 & 0.967 & 0.958 & 0.970 & 0.926 & \textbf{0.980} \\
 & CVI & 0.183 & 0.158 & 0.077 & 0.124 & 0.091 & 0.218 & \textbf{0.063} \\ \hline
\end{tabular}
\end{center}
\end{table*}

\subsection{Ablation Study}

This evaluates the effect of different components using 7 sets of experiments. \textbf{Scheme A}: without Head-Mid and Mid-Tail constraints ($\alpha^{(0,m)}$ = $\alpha^{(m,L)}$ = 0); \textbf{Scheme B}: without Mid-Tail constraint ($\alpha^{(m,L)}$ = 0); \textbf{Scheme C}: without Head-Tail constraint ($\alpha^{(0,L)}$ = 0); \textbf{Scheme D}: without Head-Tail and Mid-Tail constraints ($\alpha^{(0,L)}$ = $\alpha^{(m,L)}$ = 0); \textbf{Scheme E}: without reconstruction ($\alpha^{(0,0')}$ = 0), \textbf{Scheme F}: without three LGP constraints ($\alpha^{(0,L)}$ = $\alpha^{(m,L)}$ = $\alpha^{(m,L)}$ = 0) and \textbf{Scheme G}: the full model (baseline). The ACC/NMI/CVI metrics for the above seven schemes are reported in Table \ref{tab:ablation}.

Comparing Scheme G with the other six schemes, we can draw the following conclusions. (1) \textbf{Scheme G \textit{vs} A}: The Head-Mid constraint is very important if clustering is to be done in the clustering space, otherwise it will reduce the clustering accuracy. This may be because the Head-Mid LGP constraint allows clustering space to better preserve geometry in input data, which is conducive to better clustering results; (2) \textbf{Scheme G \textit{vs} B}: It is clear that, based on the CVI metric (especially on the USPS dataset), the Mid-Tail LGP constraint is crucial to ensure consistency between clustering and visualization; (3) \textbf{Scheme G \textit{vs} C}: The lack of Head-Tail LGP constraint leads to very poor visualization, with clusters closely adjacent to each other and with significantly more noise, especially on the MNIST-full and Fashion-MNIST dataset; (4) \textbf{Scheme G \textit{vs} D}: Good clustering can be achieved with only Head-Mid constraint on some datasets, but if no constraint is imposed on the visualization space, the visualization is poor and does not reflect the geometry of the input data at all; (5) \textbf{Scheme G \textit{vs} E}: Imposing AE reconstruction loss is beneficial to slightly improve clustering and visualization performance. (6) \textbf{Scheme G \textit{vs} F}: No proper clustering and visualization results can be obtained if none of the LGP constraints is applied. The visualization results in Fig.~\ref{fig:ablation} in \textbf{Appendix A2} are consistent with quantitative evaluation. 

\subsection{Ability to Generalize}

\begin{figure}[htbp]
	\begin{center}
		\includegraphics[width=1.0\linewidth]{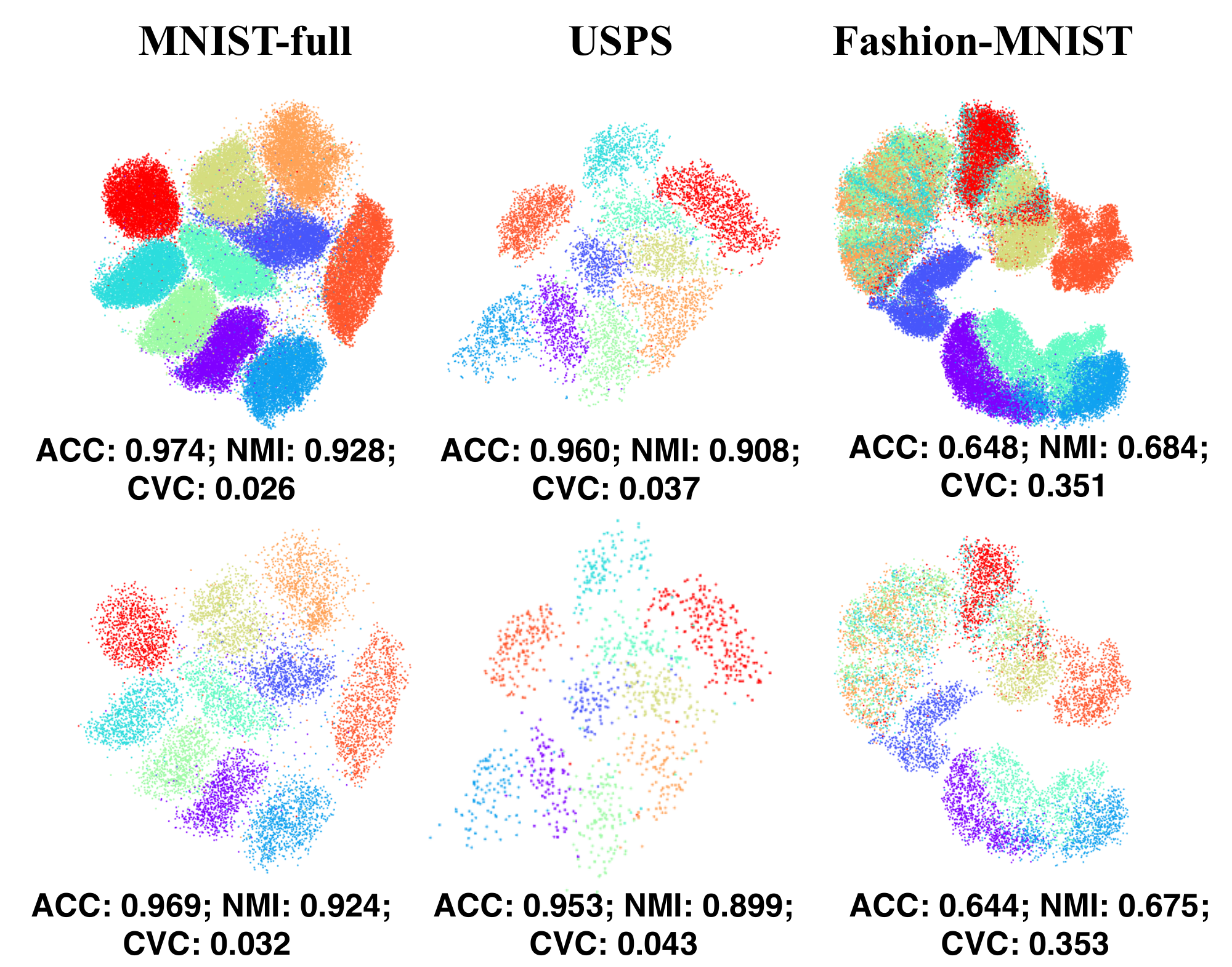}
	\end{center}
	\caption{Visualization and performance metrics on the training (upper row) and test (lower row) samples, showing excellent ability of CRL-encoder to generalize to unseen data.}
	\label{fig:generalization}
\end{figure}

Fig.~\ref{fig:generalization} demonstrates that a learned CRL-encoder network can generalize well to unseen data with high clustering accuracy. On the MNIST-full and Fashion-MNIST datasets, CRL is trained using 60,000 training samples and then tested on the remaining 10,000 test samples using the learned CRL-encoder model. For the USPS dataset, the training and test samples are 8000 and 1296, respectively. In terms of metrics, there is only a small degradation on the test samples compared to that of the training samples. The visualization also shows that CRL still maintains clear inter-cluster boundaries even on the test samples, further demonstrating the excellence of the CRL-encoder to generalize.

\vspace{-3mm}
\section{Conclusion}



The CRL method improves the consistencies among the associated tasks of NLDR, clustering, and visualization of high dimensional data in a single neural network. The CRL network applies two NLDR transformations in cascade for clustering and visualization, under the LGP constraints for preserving the structure of data along with the two-stage NLDR transformations. It is the two-stage NLDR structure and the LGP constraints that have helped improve the consistencies. Extensive results are provided to show the advantages over the most popular nonlinear methods of t-SNE and UMAP and other similar methods. While having achieved good results, the CRL presented in this paper is an initial effort. Future work can integrate various strategies used in self-supervised learning, including CNN, data augmentation, contrastive learning, and adversarial learning, to improve representation learning, clustering, and visualization in large-scale datasets such as ImageNet and CIFAR-100.

{\small
\bibliographystyle{ieee_fullname}
\bibliography{MLDL,vision}
}

\clearpage
\section*{Appendix}

\setcounter{table}{0} 
\setcounter{figure}{0} 
\renewcommand{\thetable}{A\arabic{table}}
\renewcommand{\thefigure}{A\arabic{figure}}

\subsection*{A.1 Hyperparameters}
Table \ref{tab:hyperparameter} summarizes the hyperparameter settings for different datasets. The other hyperparameter settings, which are the same for all datasets, have been described in the paper.

\begin{table}[!htbp]
\small
\centering
\caption{Hyperparameters of CCV for different datasets}
\begin{tabular}{@{}lllllllccl@{}}
\toprule
Dataset         & Point      & Dimension      & Q       & Batchsize \\ \midrule
MNIST-test      & 10000      & 784            & 15      & 5000 \\
MNIST-full      & 70000      & 784            & 20      & 4000 \\
USPS            & 9296       & 256            & 5       & 1000 \\
Fashion-MNIST   & 70000      & 784            & 20      & 4000 \\
Coil-100        & 7200       & 49152          & 3       & 2400 \\
\bottomrule
\end{tabular}
\label{tab:hyperparameter}
\end{table}

\begin{figure*}[bp]
    \begin{center}
        \includegraphics[width=1.0\linewidth]{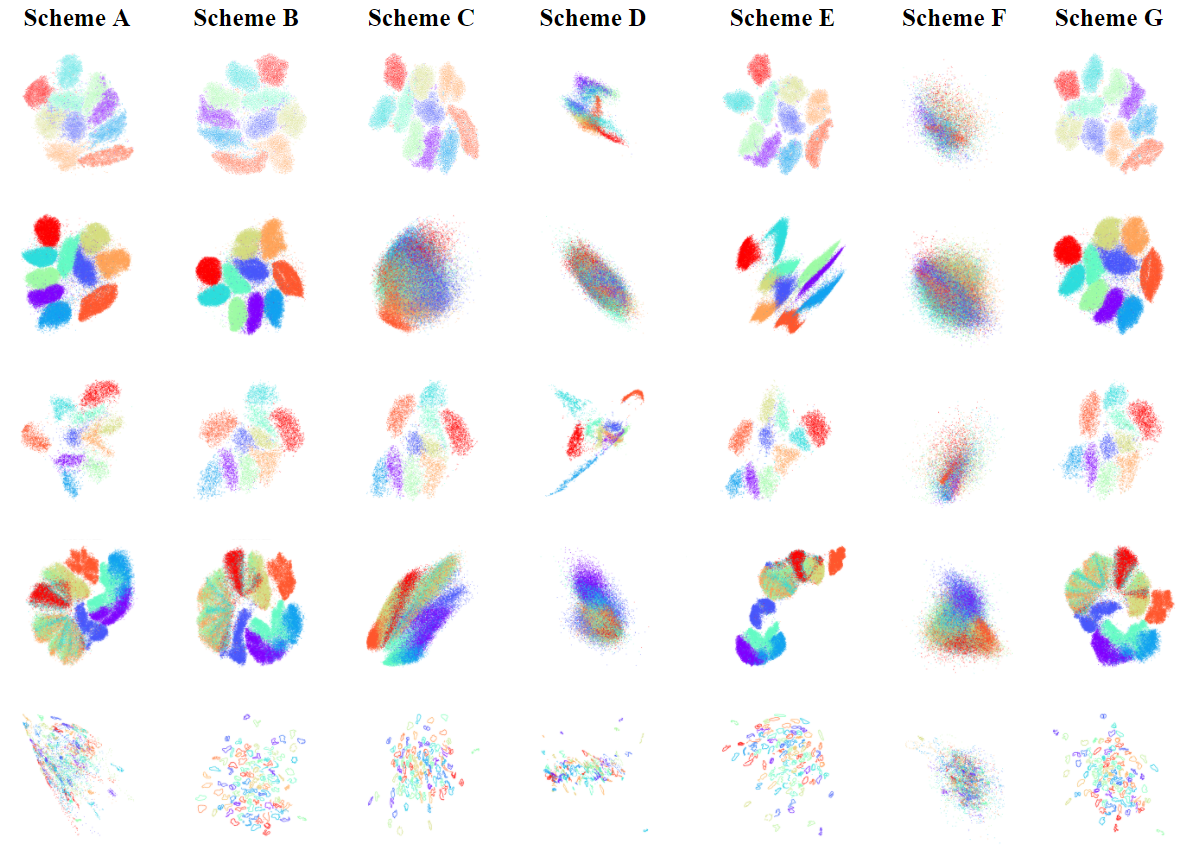}
    \end{center}
    \caption{Comparison of visualization results for ablation study. From top to bottom, by row are the visualizations on the MNIST-test, MNIST-full, USPS, Fashi-MNIST, and Coil-100 datasets, respectively.}
    \label{fig:ablation}
\end{figure*}

\subsection*{A.2 Visualization for ablation study}
The visualizations of the ablation study are shown in Fig.~\ref{fig:ablation}, along with the Table 4 in the paper to analyze the components of the CRL framework. In terms of visualization, the effects of Scheme C, D, and F are very poor, mainly due to the lack of Head-Tail constraint, which makes the visualization results inconsistent with the geometry of the input. The lack of reconstruction loss also affects the visualization to some extent, especially on the MNIST-full and Fashion-MNIST datasets.

\end{document}